\documentclass[letterpaper]{article} 
\usepackage{aaai24}  
\usepackage{times}  
\usepackage{helvet}  
\usepackage{courier}  
\usepackage[hyphens]{url}  
\usepackage{graphicx} 
\usepackage{natbib}  
\usepackage{caption} 
\usepackage{algorithm}
\usepackage{algorithmic}
\usepackage{amsmath}
\usepackage{amssymb}
\usepackage{booktabs}
\usepackage{multicol}
\usepackage{dsfont}
\usepackage{multirow}

\usepackage{newfloat}
\usepackage{listings}
\DeclareCaptionStyle{ruled}{labelfont=normalfont,labelsep=colon,strut=off} 
\floatstyle{ruled}
\newfloat{listing}{tb}{lst}{}
\floatname{listing}{Listing}
\title{Mining Gaze for Contrastive Learning toward Computer-Assisted Diagnosis}
\author{
    Zihao Zhao\textsuperscript{\rm 1}\equalcontrib,
    Sheng Wang\textsuperscript{\rm 1,\rm 2,\rm 3}\equalcontrib,
    Qian Wang\textsuperscript{\rm 1,\rm 4},
    Dinggang Shen\textsuperscript{\rm 1,\rm 3,\rm 4}\thanks{Corresponding author.}
}
\affiliations{
    \textsuperscript{\rm 1}School of Biomedical Engineering \& State Key Laboratory of Advanced Medical Materials and Devices,\\ ShanghaiTech University, Shanghai, China\\
    \textsuperscript{\rm 2}School of Biomedical Engineering, Shanghai Jiao Tong University, Shanghai, China\\
    \textsuperscript{\rm 3}Shanghai United Imaging Intelligence Co., Ltd., Shanghai, China\\
    \textsuperscript{\rm 4}Shanghai Clinical Research and Trial Center, Shanghai, China\\

    \{zhaozh2022, qianwang, dgshen\}@shanghaitech.edu.cn, wsheng@sjtu.edu.cn
}

\begin{document}

\maketitle

\begin{abstract}
Obtaining large-scale radiology reports can be difficult for medical images due to various reasons, limiting the effectiveness of contrastive pre-training in the medical image domain and underscoring the need for alternative methods. In this paper, we propose eye-tracking as an alternative to text reports, as it allows for the passive collection of gaze signals without disturbing radiologist's routine diagnosis process.
By tracking the gaze of radiologists as they read and diagnose medical images, we can understand their visual attention and clinical reasoning. When a radiologist has similar gazes for two medical images, it may indicate semantic similarity for diagnosis, and these images should be treated as positive pairs when pre-training a computer-assisted diagnosis (CAD) network through contrastive learning.
Accordingly, we introduce the Medical contrastive Gaze Image Pre-training (McGIP) as a plug-and-play module for contrastive learning frameworks. McGIP uses radiologist's gaze to guide contrastive pre-training. We evaluate our method using two representative types of medical images and two common types of gaze data. The experimental results demonstrate the practicality of McGIP, indicating its high potential for various clinical scenarios and applications.
\end{abstract}

\section{Introduction}
\label{sec:intro}

Gaze is a rich bio-signal that provides information on where an individual's eyes are directed. The collection of gaze data has significantly advanced in recent years in terms of ease, cost (typically under 300 USD), and speed~\cite{elfares2023federated,uppal2023decoding,valliappan2020accelerating,wan2021pupil}.
Gaze data has been extensively researched and utilized across multiple fields, such as marketing~\cite{wedel2017review}, robotics~\cite{aronson2022gaze,aronson2021inferring,palinko2016robot,biswas2022mitigating,manzi2020understanding}, virtual reality~\cite{hu2019sgaze,hu2021ehtask,matthews2020rendering,hu2021fixationnet,hu2020gaze}. 
In addition to these fields, eye tracking has also gained attention in the medical imaging domain as a low-cost and convenient tool~\cite{wang2022follow,ma2023eye,karargyris2021creation}. For example, the Gaze Meets ML Workshop, endorsed by MICCAI, was held in 2022 to explore the application of gaze data in medical image analysis~\citep{gazemeetml}.
One advantage of eye tracker in clinical medical imaging settings is that they do not require radiologists to open additional software programs or tools. Instead, the eye tracker can be easily attached beneath the monitor, allowing the radiologist to continue working with their existing tools and software. This can save time and effort compared to traditional annotation tools like drawing masks, circles, or boxes, which require radiologists to switch between different programs and interfaces. Moreover, eye tracking can provide additional data that cannot be captured by traditional annotation tools, such as insight into the radiologist's attentional processes and decision-making strategies.

\begin{figure}[t]
  \centering
   \includegraphics[width=0.48\textwidth]{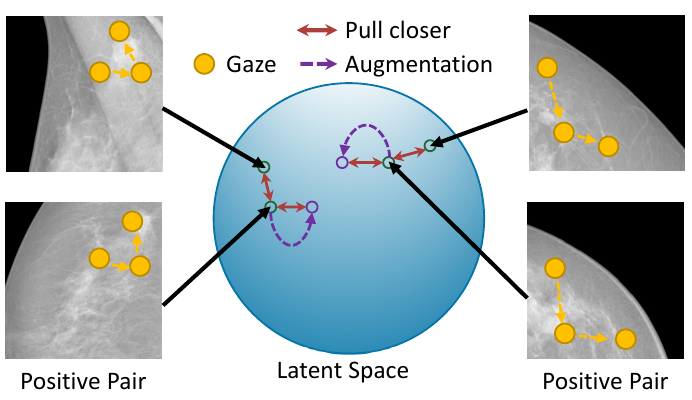}
   \caption{
   For contrastive pre-training, positive pairs are often only constructed between an image and its augmented version. In our McGIP, the images with similar gaze patterns when read by a radiologist are also considered as positive pairs and be pulled closer in the latent space.}
   \label{fig:overview}
\end{figure}

\begin{figure*}[t]
  \centering
   \includegraphics[width=0.98\textwidth]{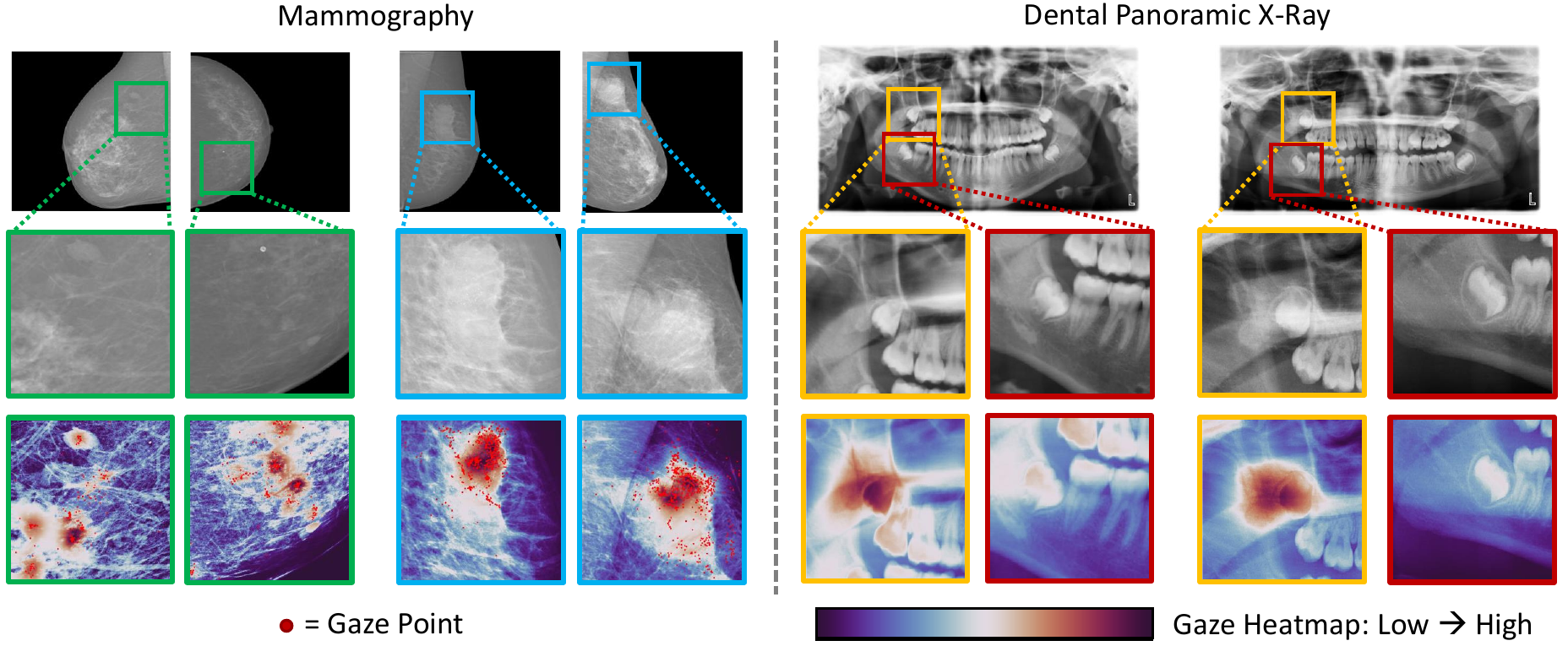}
   \caption{We show examples, with images of similar semantics corresponding to similar gaze patterns. On the left, there are four breast mammographic images, among which two are benign masses (green boxes) and two are malignant masses (blue boxes). The distributions of gaze points are similar across two benign masses, and also across two malign masses. On the right, there are two dental X-ray images of different patients. The yellow and red boxes indicate wisdom teeth on the upper and lower jaws, respectively. Across two images, the teeth of the same location have similar gaze heatmaps, corresponding to shared anatomical roles and the underlying image semantics.}
   \label{fig:similar_gaze}
\end{figure*}
In order to utilize these information in gaze, contrastive learning is a natural framework to choose since it has already successfully mine information in many cross modality data~\cite{radford2021learning}. In contrastive pre-training, images sharing similar semantics should be considered positive pairs, and vice versa. Conventional approaches~\cite{chen2020simple} create positive pair by randomly augmenting an image twice as illustrated in Figure~\ref{fig:overview}.
1) One straightforward improvement to generate better positive pairs is to use radiological reports that describe lesion locations~\cite{wiehe2022language,seibold2022breaking,vu2021medaug}. However, accessing a large set of reports is not always easy~\cite{johnson2019mimic}.
2) Another option is to use classification labels, which are more commonly available. However, generating positive pairs based on these labels presents a problem: since medical images have very few classes, there will be too many positive pairs in a contrastive batch, resulting in a collapsed representation~\cite{grill2020bootstrap}.
3) Moreover, classification labels in medical image analysis, such as BI-RADS~\cite{LauraLiberman2002BreastIR}, reflect severity rather than visual pattern. For instance, two BI-RADS 3 mammography may have vastly different lesions, i.e., small calcification and large mass. Positive pair of these two images will not lead to good representation.

In this work, we propose a strategy, \textbf{M}edical \textbf{c}ontrastive \textbf{G}aze \textbf{I}mage \textbf{P}re-training (McGIP), to use radiologist's gaze to generate additional positive pairs for medical images. As a substitute for radiological reports or diagnosis label, gaze data are 1) easy to access, 2) highly variant and 3) directly related to visual pattern of each lesion.
Originating from observations during gaze collection, we found that \textbf{medical images corresponding to similar gaze patterns, when read by a radiologist, are often positively paired and thus should be drawn closer in the latent space}, as illustrated by Figure~\ref{fig:overview}.
Specifically, \textbf{a} radiologist delivered similar gaze patterns when presented with medical images of the same semantical type. 
Two examples in Figure \ref{fig:similar_gaze} illustrate our observation.

On the left of Figure \ref{fig:similar_gaze}, we show four examples of breast mammography corresponding to benign (in green boxes) and malignant (in blue boxes) masses, respectively.
When looking at the exemplar benign masses in green boxes (BI-RADS 3, with cancer probability less than 2\%), the radiologist often shows a more ``scattered" pattern for the distribution of the gaze points.
In contrast, when looking at malignant masses (BI-RADS 4C, with cancer probability 50-95\%, and BI-RADS 5, with cancer probability at least 95\%), radiologist tends to have a much more ``centered" gaze pattern.
On the right of Figure~\ref{fig:similar_gaze}, we show two dental panoramic X-ray images.
When zooming into  \textit{tooth \#1} (denoted with yellow box),
one can notice that the gaze heatmaps are always similar when radiologist views images from two different patients. 
Meanwhile, the gaze heatmaps of \textit{tooth \#32} (denoted with red boxes) are also similarly low in magnitude across patients, implying little attention devoted to the molars from the radiologist. 
Other clinical researches also report that different types of abnormality can lead to different gaze patterns~\cite{kundel2008using, voisin2013investigating}.

We design different schemes to properly measure gaze similarity under different conditions, so that our proposed method can be generalized to various types of medical images.
In summary, our main contributions are as follows.
\begin{itemize}
\item To the best of our knowledge, McGIP is the first work to utilize human gaze as an alternative role of medical reports to guide contrastive pre-training.

\item We investigate three schemes of gaze similarity evaluation, to serve different types of medical images and also representations of gaze data.

\item We validate McGIP on two very different medical image diagnosis tasks of breast mammography and dental panoramic X-rays. The performance shows its effectiveness and generalizability for potential clinical applications.
\end{itemize}
The code implementation of our method is released at https://github.com/zhaozh10/McGIP.
\section{Related Works}

\label{sec:related_works}
In this section we first introduce gaze and its application in radiology. Then we briefly cover the topic of contrastive learning and corresponding positive pair selection.

\subsection{Gaze in Radiology}

Visual attention has proven a useful tool to understand and interpret radiologist's reasoning and clinical decision. 
 ~\citet{carmody1981finding} published one of the first eye-tracking studies in the field of radiology, where they studied the detection of lung nodules in chest X-ray films. 
In mammography, a strong correlation is found between gaze patterns and lesion detection performance \cite{kundel2008using,voisin2013investigating}.

Recently, studies start to investigate the potential of gaze in medical image analysis from the deep learning perspective. 
~\citet{mall2018modeling} and~\citet{mall2019missed} investigated the relationship between human visual attention and CNN in finding missing cancer in mammography. 
~\citet{karargyris2021creation} developed a dataset of chest X-ray images and gaze coordinates. They used a multi-task framework to perform classification for diagnosis and prediction of the gaze heatmap from radiologists at the same time.
~\citet{wang2022follow} proposed Gaze-Attention Net, which used gaze as extra supervision other than only ground-truth labels.

\subsection{Contrastive Learning}
Large-scale contrastive pre-training has become popular due to its generalizability to many scenarios and robustness against overfitting~\cite{radford2021learning}.
There are several attempts to utilize contrastive pre-training in medical image analysis.~\citet{sowrirajan2021moco} proposed MoCo-CXR to produce the models with better representations and initialization for detection of abnormalities in chest X-rays. 
~\citet{azizi2021big} utilized multiple images of the underlying pathology per patient to construct more informative positive pairs for multi-instance contrastive learning. 
These works have adopted image-augmentation-based semantic-unaware strategies to generate positive pairs. 

In the early days of contrastive learning, good representation requires a large number of negative pairs in a batch~\cite{chen2020simple,he2020momentum}. 
However, more recently, negative pairs are shown to be less necessary for learning a good representation. That is, the number of negative pairs may have a limited influence on the representation quality when the framework is designed properly~\cite{caron2021emerging}. 
In this paper, the roles of positive pairs and their impact on the learned representations will also be our focus.

While it is critical to design positive pairs in contrastive learning, most existing frameworks apply semantic-unaware data augmentation that is adapted from the conventional supervised learning~\cite{he2020momentum,chen2020simple,grill2020bootstrap}. Recent studies have found that a semantic-aware contrastive learning process can perform better.
~\citet{selvaraju2021casting} proposed CAST to use unsupervised saliency maps to sample the crops. ~\citet{peng2022crafting} proposed ContrastiveCrop for augmentation of semantic-aware cropping.

\section{Method}
\begin{figure*}[htbp]
  \centering
   \includegraphics[width=1\linewidth]{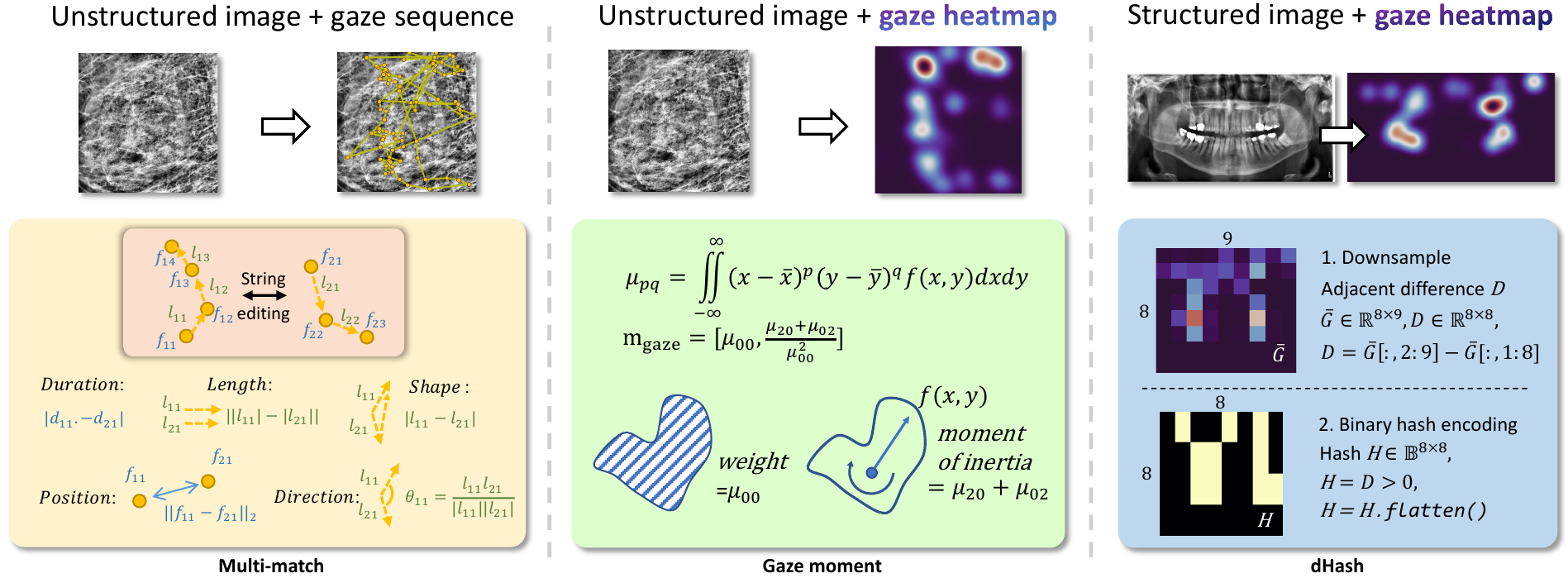}
  \caption{Three schemes to compute gaze similarity for different image types and gaze representations. 
  On the left, for unstructured images, we extract five features for each gaze sequence, and calculate the inter-sequence similarity by the multi-match algorithm~\cite{multimatch}. 
  In the middle, also for unstructured images, we use heatmap as gaze representation, and calculate the similarity by gaze moment. 
  On the right, for structured images, we dHash each heatmap into an $8\times8$ code.}
   \label{fig:framework}
\end{figure*}

This section first discusses the collection and processing of gaze signals. Then, we propose different gaze similarity evaluation schemes for structured and unstructured medical images. Finally, we use gaze similarity to appropriately generate positive pairs and integrate them with multiple contrastive pre-training frameworks.

\subsection{Gaze Collection and Processing}
Gaze collection can be a seamless process when radiologists conduct routine diagnoses. 
Specifically, we used a Tobii pro nano remote eye-tracker to collect binocular gaze data at 60Hz. A radiologist with ten years of experience was invited to read and diagnose images on a computer, e.g., from a breast mammography dataset. 
A graphic user interface was developed to adapt to the typical clinical workflow, offering functions of multi-window viewing and interactive operations such as zooming, contrasting, etc. 
In this way, the gaze data can be collected from a nearly real diagnosis environment for the radiologist, which reduces interference during collection~\cite{ma2023eye}.

The recorded eye-tracking data consists of a temporal sequence of points, each of which comes with the location and the timestamp. 
The gaze points need to be further categorized into saccade points (fast eye movement) and fixation points~(yellow dots in Figure~\ref{fig:framework}, denoted as $f_{ij}$ for multi-match algorithm~\cite{multimatch}).
Specifically, the centroid of a small cluster of the raw gaze points is considered as a fixation point, with the time-lapse of the cluster as its duration.
Note that the saccade points indicate rapid eye movement, which corresponds to object searching in a large field-of-view for human vision system. Thus, the saccade points carry global features of the images. 
In contrast, the locations and duration of fixation points, corresponding to radiologist's visual focus on specific regions in the image, can reveal local features. 
A patch containing normal tissue only may have fewer than ten fixation points, while a lesion patch can have much more.
The gaze points can be expressed by either gaze sequence (preserving both temporal and spatial information of the gaze) or gaze heatmap (providing spatial distribution only).

\subsection{Gaze Similarity Evaluation for Different Scenarios}
\label{gaze_similarity}
It is pivotal to evaluate gaze similarity in our work, as images with similar gaze patterns are presumably close to each other in semantics. 
We have divided medical images into two categories roughly, namely \textbf{structured} and \textbf{unstructured} images. 
In structured images, the patients are typically well positioned and imaged following strict clinical protocols, and radiologist's gaze tends to be similar in the same regions across different images. The reason is that those regions typically correspond to the same anatomic structure as illustrated by the example in the right of Figure~\ref{fig:similar_gaze}.
The unstructured images, in comparison, are very different -- the patients are not imaged with predefined position, and the images, e.g. mammography and pathological image, are usually interpreted at patch-level. In this case, the global anatomical structure is not a major clue to support the diagnosis, as shown in the left of Figure~\ref{fig:similar_gaze}.

For a batch of images $[x_1, x_2,...,x_n]$, we assume ${A}$ affinity matrix for gaze similarity.
In this work, we investigate three different ways to evaluate gaze similarity for different categories of medical images. 
For the unstructured images, we utilize both gaze sequence (in the form of $[time, location]$ for each fixation point) and gaze heatmap, as the two data formats are both commonly adopted. And for the structured images, we propose to evaluate the gaze similarity by referring to the gaze heatmap.

\textbf{Unstructured Image + Gaze Sequence.}
\citet{holmqvist2011eye-tracking-guide} described gaze sequences from five different features, i.e. shape, length, direction, position and duration. 
We further calculate the inter-sequence gaze similarity from these five features based on multi-match \cite{multimatch}.
That is, the comparison between two gaze sequences of varying lengths is considered as a string-editing problem, where the minimum editing cost serves the dissimilarity.

Assuming two gaze sequences $G_1$ and $G_2$ contain 3 and 4 points as shown in Figure \ref{fig:framework}. $f_{11},f_{21}$ are two fixation points in $G_1$ and $G_2$, $l_{11},l_{21}$ are offsets between two consecutive fixations, and $d_{11}, d_{21}$ indicate durations of $f_{11}$ and $f_{21}$.
The inter-point duration editing cost is the relative duration difference, 
i.e. ${|d_{11}-d_{21}|}/{\max(d_{11},d_{21})}$.
We thus construct $S_{dur}\in \mathbb{R}^{4\times3}$ based on pairwise duration editing cost. 
The minimum \textit{duration} editing cost is the minimum travel cost from the top left of $S_{dur}$ to the bottom right, which can be formulated as a classic dynamic programming problem.
Analogously, the \textit{position}, \textit{shape}, \textit{length} and \textit{direction} costs can also be computed through corresponding editing costs as shown in Figure \ref{fig:framework}.
The multi-match shall first measure the similarity between two gaze sequences from the five above-mentioned aspects. The overall similarity, denoted by $A_{12}$, is then calculated by weighted summation. Similarly, the $A_{ij}$ of gaze sequences $G_i$ and $G_j$ is also computed using the same approach.

\textbf{Unstructured Image + Gaze Heatmap.}
An even more common way to represent gaze data is to use the heatmap, which is generated by convoluting the raw gaze points with Gaussian filters~\cite{le2013methods}. Based on the heatmaps, we adopt a classic method of image moment to measure their similarity.
In general, the family of moments is defined as
\begin{equation}
    \mu_{pq}=\iint\limits_{-\infty}^{\quad\infty}(x-\bar{x})^{p}(y-\bar{y})^{q}f\left(x,y\right)dxdy,
\end{equation}
where $f(x,y)$ is gaze heatmap, $(\bar{x}, \bar{y})$ is the centroid of $f(x,y)$ and $p,q \in \mathbb{N}$ define the order of moment. 
Then, the first invariant of Hu-moment~\cite{hu1962visual} is adopted here to measure the dispersion of the heatmap in both row and column directions:
\begin{equation}
    \phi_1=\frac{\mu_{20}+\mu_{02}}{\mu_{00}^2},
    \label{hu_moment}
\end{equation}
where $\mu_{20}+\mu_{02}$ is the moment of \textit{inertia}. We also use $\mu_{00}$, which is the \textit{weight} moment and duration in our case. 
Thus, the gaze is described by its moment vector $m_{gaze}=[\mu_{00},\phi_1]$. Beware that gaze moment is introduced here to measure the difference. In this case, the affinity between two moment vectors $m_{gaze}^i$ and $m_{gaze}^j$ is defined as
\begin{equation}
\begin{aligned}
    A_{ij}=& \alpha(1-\delta(\mu_{00}^i,\mu_{00}^j))\\
    +&(1-\alpha)(1-\delta(\phi_{1}^i,\phi_{1}^j)),
\end{aligned}
    \label{diff2sim}
\end{equation}
where the affinity is defined as one minus the normalized $L_1$ distance between two moment vectors: $\delta(x,y)=\frac{L_1(x,y)}{\textrm{max}(x,y)}$, and $\alpha$ is a manually selected hyper-parameter. We set $\alpha$=0.5 in our experiments. 

\textbf{Structured Image + Gaze Heatmap.}
In this circumstance, we adopt dHash~\cite{imagehash_survey}, a widely-used image matching algorithm, to embed a heatmap into 64-bit hash code and then measure the similarity. 
As in Figure~\ref{fig:framework}, dHash first resizes the gaze heatmap $G$ into $\bar{G}\in \mathbb{R}^{8\times 9}$, thus filtering out much high-frequency components yet preserving lasting fixations. In each row, we compute the difference between adjacent pixels and get $D\in\mathbb{R}^{8\times 8}$. Here, the computation happens in the row direction, because the teeth in our exemplar dataset are aligned in this way.
The binary mask $H\in \mathbb{B}^{8\times 8}$ is then encoded by thresholding $D>0$. The similarity $A_{ij}$ between two heatmaps can thus be measured as cosine similarity by flattening $H$'s.

\subsection{Contrastive Pre-training with Gaze}
Typical contrastive learning methods usually construct one positive pair for each sample, while the proposed McGIP can construct several positive pairs for each sample in a batch. 
For a batch of images $[x_1, x_2,...,x_n]$, ${A}$ is the affinity matrix for gaze similarity.
Denote the constraint function in contrastive learning as $\textrm{CST($\cdot)$}$, which represents InfoNCE in MoCo and $L_2$ distance in BYOL for example. The overall loss function for a batch is
\begin{align}
    L=\frac{\sum_{i,j=1}^{n}  \mathds{1}_{A_{ij}\geq t}{(A_{ij})\cdot\textrm{CST}(\textrm{Net}(x_i),\textrm{Net}(\hat{x}_j))}}{\sum_{i,j=1}^{n}\mathds{1}_{A_{ij}\geq t}{(A_{ij})}},
    \label{loss_function}
\end{align}
where $\mathds{1}$ refers to indicator function, $\hat{x}_j$ denotes the transformed view of $x_j$, and ``\textrm{Net}'' indicates the encoder of contrastive learning.
The term $t$ here is a threshold to determine whether the two images are similar enough to be a positive pair.
The above Eq. \eqref{loss_function} will be the optimization objective during each iteration.
Moreover, gaze data inherently contains noise and uncertainty, which is especially significant on unstructured images. We correspondingly introduce $p$, denoting confidence score, for unstructured images. When selecting positive unstructured image pairs based on gaze heatmaps, we only consider it a true positive with a possibility of $p$ if the gaze similarity is higher than $t$.

\section{Experimental Results}
\label{sec:exp}
\subsection{Datasets and Metrics}
\begin{table*}[htbp]
  \centering
  \caption{Fine-tuning performance on the INbreast dataset of McGIP with different contrastive learning frameworks.}
  \resizebox{\textwidth}{!}{
    \begin{tabular}{cccccccccc}
    \toprule
    \multirow{2}[4]{*}{Method} & \multicolumn{3}{c}{ResNet18} & \multicolumn{3}{c}{ResNet50} & \multicolumn{3}{c}{ResNet101} \\
    \cmidrule(lr){2-4}   \cmidrule(lr){5-7}\cmidrule(lr){8-10} 
          & M-AUC & AUC & ACC & M-AUC & AUC & ACC & M-AUC & AUC & ACC \\
    \midrule
    From-scratch& 71.39±3.26 & 73.98±4.55 & 76.76±4.79 & 68.01±2.41 & 71.44±4.38 & 75.41±4.22 & 69.89±2.56 & 67.27±3.71 & 73.78±2.02 \\
    ImageNet & 83.43±1.98 & 82.38±2.84 & 80.38±2.34 & 89.73±0.89 & 86.17±1.23 & 82.97±1.08 & 88.90±2.98 & 87.50±0.89 & 85.63±2.26 \\
    \midrule
    MoCo  & 82.19±3.05 & 84.69±2.53 & 82.43±1.71 & 89.52±2.15 & 89.44±0.90 & 81.62±1.38 & 92.28±2.86 & 91.03±1.88 & 86.22±1.58 \\
    MoCo+McGIP & \textbf{85.07±2.43} & \textbf{88.37±1.73} & \textbf{83.51±1.01} & \textbf{92.74±1.87} & \textbf{91.44±2.08} & \textbf{85.68±1.38} & \textbf{93.06±1.73} & \textbf{92.58±2.92} & \textbf{87.03±0.66} \\
    \midrule
    BYOL  & 90.42±2.31 & 90.59±1.48 & 83.78±0.85 & 93.84±1.72 & 87.96±1.71 & 85.95±1.83 & 93.82±3.44 & \textbf{90.39±2.08} & 86.49±0.89 \\
    BYOL+McGIP & \textbf{95.83±0.63} & \textbf{94.96±1.13} & \textbf{85.14±0.85} & \textbf{97.07±0.75} & \textbf{93.80±0.79} & \textbf{87.57±1.01} & \textbf{95.46±2.67} & {90.09±3.08} & \textbf{86.76±0.54} \\
    \midrule
    SimSiam & 91.10±3.26 & 91.81±1.63 & 83.51±1.01 & 93.11±2.26 & 86.56±2.99 & 86.27±1.79 & 92.26±1.11 & \textbf{90.26±1.14} & 85.68±1.08 \\
    SimSiam+McGIP & \textbf{95.30±1.16} & \textbf{94.62±1.34} & \textbf{85.95±1.38} & \textbf{95.30±0.78} & \textbf{89.22±0.95} & \textbf{88.65±1.38} & \textbf{96.85±0.63} & {90.08±1.51} & \textbf{87.84±1.01} \\
    \bottomrule
    \end{tabular}}%
  \label{tab:finetune_INBreast}%
\end{table*}%
\begin{table}[htbp]
  \centering
  \caption{Fine-tuning performance on the Tufts dataset of McGIP with different backbones.}
  \resizebox{0.35\textwidth}{!}{
    \begin{tabular}{cccc}
    \toprule
    Backbone & Method & AUC & ACC \\\midrule
    \multirow{4}{*}{ResNet18} & From-scratch & 47.58  & 61.00 \\
    & ImageNet & 60.26  & 60.50 \\
          & BYOL  & 60.61    & 63.00 \\
          & BYOL+McGIP & \textbf{62.91}  & \textbf{65.00} \\\midrule
    \multirow{4}{*}{ResNet50} &From-scratch & 55.30  & 61.00 \\
    &ImageNet & 60.06    & 62.00 \\
          & BYOL  & 52.12  & 63.50 \\
          & BYOL+McGIP & \textbf{61.35}  & \textbf{67.50} \\\midrule
    \multirow{4}{*}{ResNet101}  &From-scratch & 57.61  & 61.00 \\
    &ImageNet & 59.96  & 61.50 \\
          & BYOL  & 58.05    & 63.00 \\
          & BYOL+McGIP & \textbf{61.14}  & \textbf{64.50} \\\bottomrule
    \end{tabular}
}
  \label{tab:finetune_tufts}
\end{table}

We conduct experiments on two datasets: INbreast~\cite{InsMoreira2012INbreastTA} and Tufts dental dataset~\cite{panetta2021tufts}. 
 
\textbf{The INbreast dataset}~\cite{InsMoreira2012INbreastTA} includes 410 full-ﬁeld digital mammography images
collected from 115 patients. 
We invited a radiologist with 10 years of experience to diagnose the images in this dataset, and collected the eye-movement data. 
The diagnosis was following BI-RADS assessment of masses~\cite{LauraLiberman2002BreastIR}, and classified all images into three groups: normal~(302 images), benign~(37) and malignant~(71), respectively.
The gaze data was collected using a Tobii pro nano eye-tracker, and pre-processed with the toolbox proposed in~\citet{ma2023eye}.
We randomly split the INbreast dataset for five-fold cross-validation, with 80\% images for training and 20\% for test. 

To inspect the performance, we report the accuracy (ACC), area-under-curve for malignant masses (M-AUC) and area-under-curve for all three classes (AUC) on the testing data. 
Here M-AUC is specially calculated since malignant masses are critical to diagnose, whose risks to cancer can be more than 10 times higher than benign masses~\cite{LauraLiberman2002BreastIR}.
We can use multi-match and gaze moment to calculate the gaze similarity on this dataset, respectively. 
The threshold of gaze similarity is set to 0.7. For gaze moment, \textit{p} is set to 0.5 in all implemented experiments.

\textbf{The Tufts dataset}~\cite{panetta2021tufts} is composed of 1000 panoramic dental X-ray images, together with processed gaze heatmaps. 
There are two groups of images in the dataset: normal~(340) and abnormal~(660), respectively. 
We choose 70\% and 10\% of images for training and validation, while the remaining 20\% of images constitute the testing set. For the Tufts dataset, we report the accuracy (ACC) and area-under-curve (AUC) on the testing data as the performance indicators.
We use dHash to calculate the gaze similarity on this dataset and set the threshold to 0.7. 

All the experiments are implemented with PyTorch 1.13.0 on a single NVIDIA RTX3060. Unless otherwise specified, all networks are trained for 200 epochs using Adam optimizer with the learning rate (\textit{lr}) set to $2e^{-5}$ in the pre-training. Fine-tuning and linear probing for final classification are trained for 10 epochs (INbreast) and 20 epochs (Tufts) with Adam optimizer (\textit{lr}: $2e^{-5}$). All contrastive learning methods are initialized from ImageNet pre-trained weights.

\subsection{Performance on Diagnosis Tasks}
In order to demonstrate the generalizability and practicality of McGIP, we test different diagnosis tasks on the two datasets. 
In particular, we evaluate the fine-tuning performance, which is popularly adopted in medical image studies~\cite{azizi2021big}. 

The results of INbreast dataset are reported at Table~\ref{tab:finetune_INBreast}. 
Compared to the conventional way to supervise the pre-training with ImageNet-1K, the performance of McGIP improves constantly.
Moreover, compared to existing contrastive learning such as MoCo~v2~\cite{sowrirajan2021moco}~(denoted as MoCo), McGIP constantly improves with different backbones in ACC (from 83.42\% to 85.41\% avergaed over three backbones).
The same trend is also observed for other evaluation metrics, such as AUC and M-AUC, in most cases. Notably, although the AUC of McGIP is slightly lower when using the ResNet101 backbone, it demonstrates a significant advantage in terms of M-AUC, which is a more critical metric for accurate breast cancer diagnosis.
Similarly, for the classification task related to the panoramic X-ray image dataset, we report the results for only BYOL with three backbones in Table~\ref{tab:finetune_tufts} due to page limit. 
Still, McGIP constantly offers the best fine-tuning performance among all compared settings.
In summary, McGIP effectively improves contrastive learning in the final diagnosis performance, while our method is notably agnostic to different network backbones.

\subsection{Representation Quality Analysis}
\label{exp:linear}
\begin{figure*}[t]
  \centering
  \includegraphics[width=0.97\textwidth]{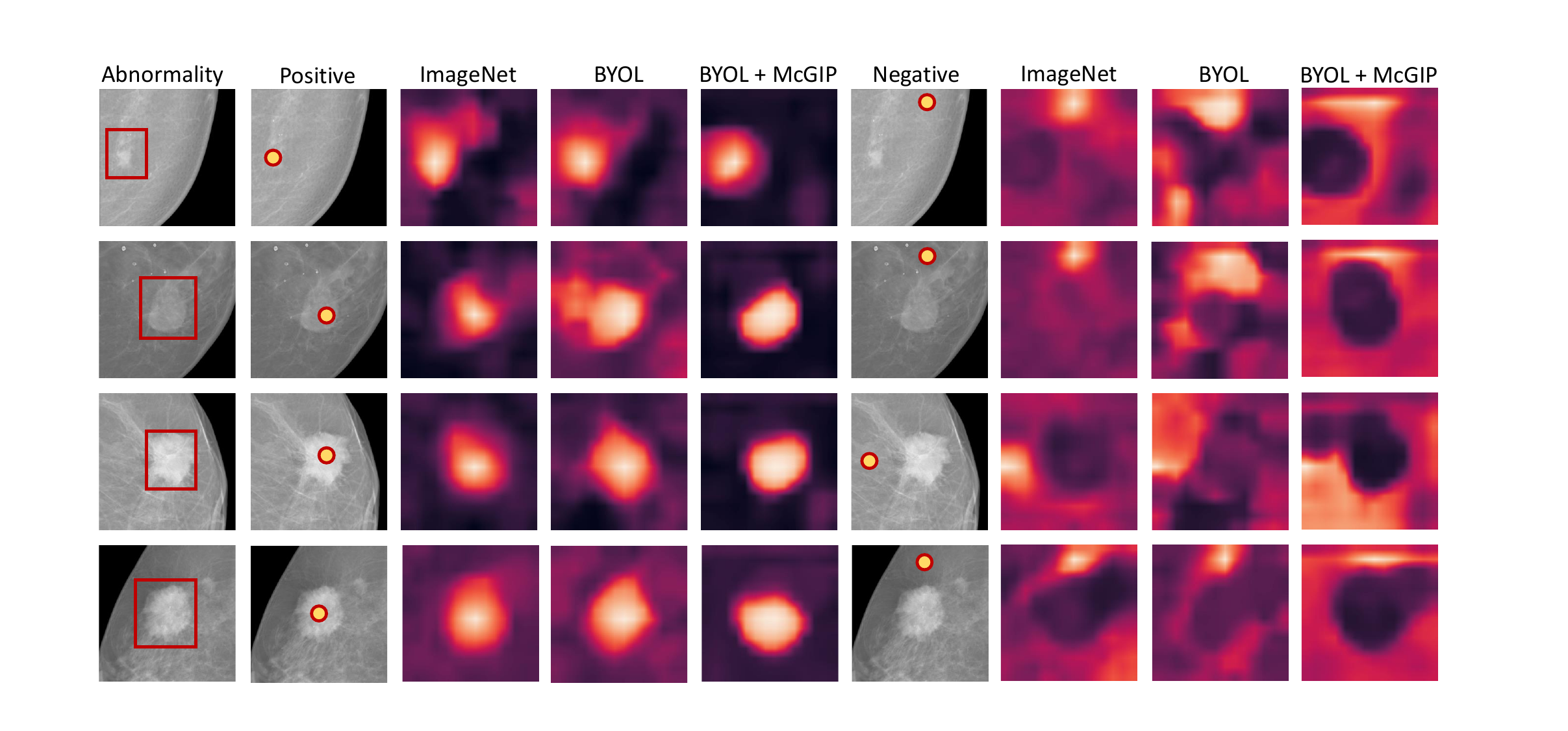}
  \caption{\textbf{Image affinity analysis.} We selected a point from feature map and calculated similarity with all other points. Natural image supervised pretrained, semantic-unaware pretrained, and McGIP pretrained weights, are used to illustrate with both positive points (abnormality) and negative points (non-abnormality). Brighter color denotes more similar points.}
  \label{fig:image_affinity}
\end{figure*}

While previous results confirm that McGIP can effectively boost classification performance with various contrastive learning frameworks, it is more interesting to inspect representation quality after contrastive pre-training. 
We visualize the point-to-point affinity in Figure~\ref{fig:image_affinity}, which highlights the quality of the learned representation of McGIP. 
Specifically, several image patches are randomly selected from the testing set of INbreast and then resized to $224\times224$.
Based on the pre-trained backbone of ResNet50, we use linear-probing weights and derive the high-resolution feature map for each patch.
Then, we randomly select a point inside the malignant mass (e.g., marked by a \textit{positive} dot in Figure~\ref{fig:image_affinity}), and calculate its affinity with all other points in the patch by cosine similarity of their corresponding feature vectors. The resulting affinity maps are shown in individual rows, while the columns are for different pre-training schemes. 
Similarly, we select \textit{negative} points from the non-mass tissues, and show the affinity maps in the right of Figure~\ref{fig:image_affinity}. 

For affinity maps of positive points, the high concentration of affinity indicates that the selected point is highly similar to nearby points located within the mass. Pre-trained models are able to encode semantic information to a considerable extent. By employing the proposed approach (BYOL+McGIP), we observe that the affinity distributions for positive points are sharper than those obtained through pre-training with ImageNet or contrastive learning using original BYOL. This observation suggests that our approach is better suited for accurately representing masses.

Our method's superiority is especially pronounced in cases where the negative points are far apart, suggesting that it could be a promising approach for classifying challenging cases in mammography. Specifically, the ImageNet pre-trained model is not good at grouping negative points that are relatively far away, as the corresponding affinity maps deliver pulse-like patterns.  Although BYOL shows slight improvements over ImageNet, it still fails to group some cases, as indicated by the last row in Figure~\ref{fig:image_affinity}. In contrast, our method exhibits significant improvements. In the last column of the figure, the boundaries between high- and low-affinity regions are clearly defined and mostly aligned with the mass contours. We attribute such superiority to the spatial semantics introduced via gaze signals, which contribute to better classification performance as evidenced by Table \ref{tab:finetune_INBreast}.

\begin{table}[htbp]
  \centering

  \caption{Performance comparison on the INbreast.}
\resizebox{0.49\textwidth}{!}{
    \begin{tabular}{ccccc}
    \hline
                                 & Method      & M-AUC      & AUC        & ACC        \\ \hline
    \multirow{2}{*}{Fine tuning}   & Gaze moment & 95.75±1.10	&93.73±1.06&	86.49±0.85 \\
                                 & Multi-match & 97.07±0.75 & 93.80±0.79 & 87.57±1.01 \\ \hline
    \multirow{2}{*}{Linear probing} & Gaze moment & 82.20±2.18	&77.11±1.54	&77.84±1.48 \\
                                 & Multi-match & 78.31±2.89 & 78.21±2.38 & 76.22±1.08 \\ \hline
    \end{tabular}
}

  \label{tab:moment-multimatch}%
\end{table}%
\subsection{Gaze Sequence vs. Gaze Heatmap}
In the method section, we introduce two approaches for measuring gaze similarity in unstructured images: \textit{multi-match} for gaze sequence and \textit{gaze moment} for gaze heatmap. The performances of these two approaches are compared in Table~\ref{tab:moment-multimatch}, based on INbreast with ResNet50. 
The two approaches perform similarly while offering various benefits. 
Gaze moment, being a renaissance variant of Hu-moment algorithm~\cite{hu1962visual}, has a straightforward and elegant form. Gaze sequence preserves more information because there are timestamps for individual spatial locations and the multi-match algorithm describes gaze similarity from multiple perspectives.
In conclusion, we recommend that users choose between these two approaches based on their specific requirements. By offering multiple options for measuring gaze similarity, we hope to enable researchers to choose the most appropriate approach for their scenarios.
In summary, we recommend that users can choose from the two approaches based on specific applications, as they have performed similarly in our experiments.
\begin{table}[htbp]
  \centering
  \caption{Comparison with supervised contrastive learning}
  \resizebox{0.5\textwidth}{!}
  {
    \begin{tabular}{ccccccc}
    \toprule
    \multicolumn{2}{c}{\multirow{2}[3]{*}{Method}} & \multicolumn{3}{c}{INbreast} & \multicolumn{2}{c}{Tufts} \\
\cmidrule(lr){3-5} \cmidrule(lr){6-7}   \multicolumn{2}{c}{} & M-AUC & AUC   & ACC   & AUC   & ACC \\
\cmidrule{1-7}
    \multirow{2}[1]{*}{ResNet18} & BYOL+Sup & 93.01±1.78 & 90.86±1.32 & 84.32±0.66 & 60.53 & 59.00 \\
          & BYOL +McGIP & \textbf{95.83±0.63} & \textbf{94.96±1.13} & \textbf{85.14±0.85} & \textbf{62.91} & \textbf{65.00} \\
    \midrule
    \multirow{2}[2]{*}{ResNet50} & BYOL+Sup & 96.02±0.88 & 88.96±1.61 & 85.14±0.85 & 59.29 & 58.00 \\
          & BYOL +McGIP & \textbf{97.07±0.75} & \textbf{93.80±0.79} & \textbf{87.57±1.01} & \textbf{61.35} & \textbf{67.5} \\
    \midrule
    \multirow{2}[2]{*}{ResNet101} & BYOL+Sup & 94.81±2.48 & \textbf{90.60±2.97} & 85.95±1.08 & 59.17 & 63.50 \\
          & BYOL+McGIP & \textbf{95.46±2.67} & 90.09±3.08 & \textbf{86.76±0.54} & \textbf{61.14} & \textbf{64.50} \\
    \bottomrule
    \end{tabular}}%
  \label{tab:gaze&gt}%
\end{table}%

\subsection{Gaze vs. Ground-Truth}
We conducted an empirical study to evaluate the effectiveness of using gaze data as a form of weak supervision compared to supervised contrastive learning, which is based on ground-truth labels. We simply consider images belonging to the same category as positive pairs and indicate it as BYOL+Sup in Table~\ref{tab:gaze&gt}.

To assess the performance of gaze data, we fine-tuned different backbones on both the INbreast and Tufts datasets. Our results show that on the INbreast dataset, gaze data outperforms ground-truth labels, albeit slightly lower in AUC when using the ResNet101 backbone. On the Tufts dataset, the advantage of gaze data is more pronounced.
In summary, our findings suggest that gaze data has greater potential than ground-truth labels and may serve as an alternative to radiological reports in the era of large models.
\begin{figure}[t]
  \centering
  \includegraphics[width=0.45\textwidth]{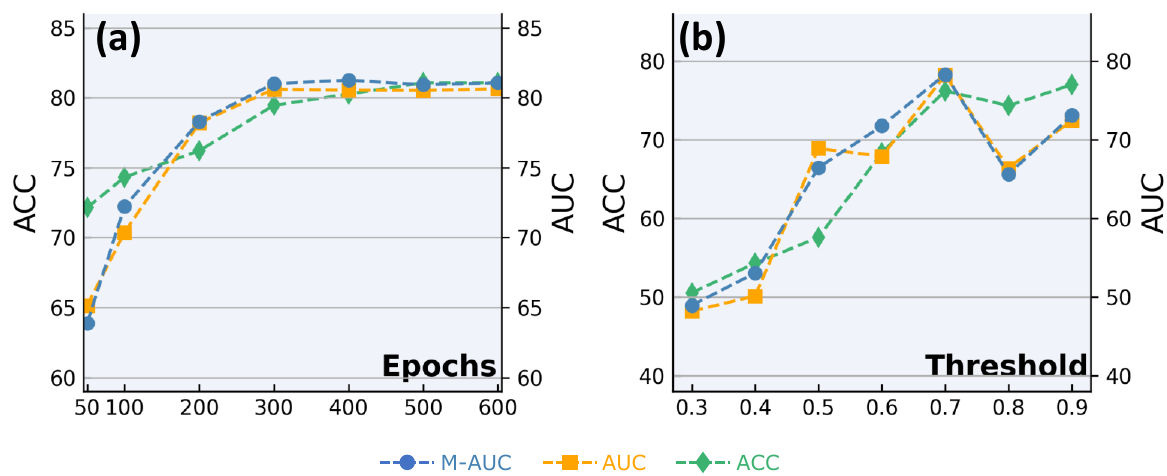}
  \caption{Linear-probing performance (backbone: ResNet50) on the INbreast dataset with (a) different pre-training epochs and (b) different similarity thresholds.}
  \label{fig:unify_ablation}
\end{figure}
\subsection{Ablation Study}
In the ablation analysis, we study how the training recipe affects the performance of the proposed McGIP. All these ablation studies are conducted under BYOL framework with a ResNet50 backbone. 
Regarding pre-training epochs, we observe a correlation between performance and numbers of epochs in Figure~\ref{fig:unify_ablation}(a). The performance gain becomes marginal when training epochs are larger than 300, which is much earlier to happen compared to natural images~\cite{chen2020simple}.
In Figure~\ref{fig:unify_ablation}(b), we report the performance of different similarity thresholds for gaze sequences. One may notice that the performance drops when the threshold is too high (i.e., 0.8 and 0.9), because McGIP has very few extra positive pairs in a mini-batch and degraded into normal contrastive learning. In contrast, when the threshold is too low (i.e., $<$0.6), there will be too many positive pairs in a mini-batch, causing a ``collapsed solution"~\cite{grill2020bootstrap}.

\section{Conclusion}
In this paper, we start with an observation that images sharing similar semantics usually have similar radiologists' gaze patterns. Therefore we explore radiologists' gazes for contrastive pre-training. 
We propose McGIP, a simple module to the existing contrastive learning, to shift more attention on the images with similar gaze patterns. 
Three schemes to evaluate gaze similarity are investigated given different medical scenarios. The superior performance on two different clinical tasks show the practicability and generalizability of McGIP, which also validate our hypothesis -- similar gaze patterns lead to similar semantics in medical images.

\appendix
\section{Acknowledgments}
This work is supported in part by The Key R\&D Program of Guangdong Province, China (grant number 2021B0101420006).

\bibliography{aaai24}

\begin{thebibliography}{45}
\providecommand{\natexlab}[1]{#1}

\bibitem[{Aronson and Admoni(2022)}]{aronson2022gaze}
Aronson, R.~M.; and Admoni, H. 2022.
\newblock Gaze complements control input for goal prediction during assisted teleoperation.
\newblock In \emph{Robotics science and systems}.

\bibitem[{Aronson, Almutlak, and Admoni(2021)}]{aronson2021inferring}
Aronson, R.~M.; Almutlak, N.; and Admoni, H. 2021.
\newblock Inferring goals with gaze during teleoperated manipulation.
\newblock In \emph{2021 IEEE/RSJ International Conference on Intelligent Robots and Systems (IROS)}, 7307--7314. IEEE.

\bibitem[{Azizi et~al.(2021)Azizi, Mustafa, Ryan, Beaver, Freyberg, Deaton, Loh, Karthikesalingam, Kornblith, Chen et~al.}]{azizi2021big}
Azizi, S.; Mustafa, B.; Ryan, F.; Beaver, Z.; Freyberg, J.; Deaton, J.; Loh, A.; Karthikesalingam, A.; Kornblith, S.; Chen, T.; et~al. 2021.
\newblock Big Self-Supervised Models Advance Medical Image Classification.
\newblock \emph{arXiv preprint arXiv:2101.05224}.

\bibitem[{Biswas et~al.(2022)Biswas, Pardhi, Chuck, Holtz, Niekum, Admoni, and Allievi}]{biswas2022mitigating}
Biswas, A.; Pardhi, B.~A.; Chuck, C.; Holtz, J.; Niekum, S.; Admoni, H.; and Allievi, A. 2022.
\newblock Mitigating causal confusion in driving agents via gaze supervision.
\newblock In \emph{Aligning Robot Representations with Humans workshop@ Conference on Robot Learning}.

\bibitem[{Carmody, Nodine, and Kundel(1981)}]{carmody1981finding}
Carmody, D.~P.; Nodine, C.~F.; and Kundel, H.~L. 1981.
\newblock Finding lung nodules with and without comparative visual scanning.
\newblock \emph{Perception \& psychophysics}, 29(6): 594--598.

\bibitem[{Caron et~al.(2021)Caron, Touvron, Misra, J{\'e}gou, Mairal, Bojanowski, and Joulin}]{caron2021emerging}
Caron, M.; Touvron, H.; Misra, I.; J{\'e}gou, H.; Mairal, J.; Bojanowski, P.; and Joulin, A. 2021.
\newblock Emerging properties in self-supervised vision transformers.
\newblock In \emph{Proceedings of the IEEE/CVF International Conference on Computer Vision}, 9650--9660.

\bibitem[{Chen et~al.(2020)Chen, Kornblith, Norouzi, and Hinton}]{chen2020simple}
Chen, T.; Kornblith, S.; Norouzi, M.; and Hinton, G. 2020.
\newblock A simple framework for contrastive learning of visual representations.
\newblock In \emph{International conference on machine learning}, 1597--1607. PMLR.

\bibitem[{Dewhurst et~al.(2012)Dewhurst, Nystr{\"o}m, Jarodzka, Foulsham, Johansson, and Holmqvist}]{multimatch}
Dewhurst, R.; Nystr{\"o}m, M.; Jarodzka, H.; Foulsham, T.; Johansson, R.; and Holmqvist, K. 2012.
\newblock It depends on how you look at it: Scanpath comparison in multiple dimensions with MultiMatch, a vector-based approach.
\newblock \emph{Behavior research methods}, 44(4): 1079--1100.

\bibitem[{Elfares et~al.(2023)Elfares, Hu, Reisert, Bulling, and K{\"u}sters}]{elfares2023federated}
Elfares, M.; Hu, Z.; Reisert, P.; Bulling, A.; and K{\"u}sters, R. 2023.
\newblock Federated Learning for Appearance-based Gaze Estimation in the Wild.
\newblock In \emph{Annual Conference on Neural Information Processing Systems}, 20--36. PMLR.

\bibitem[{Grill et~al.(2020)Grill, Strub, Altch{\'e}, Tallec, Richemond, Buchatskaya, Doersch, Pires, Guo, Azar et~al.}]{grill2020bootstrap}
Grill, J.-B.; Strub, F.; Altch{\'e}, F.; Tallec, C.; Richemond, P.~H.; Buchatskaya, E.; Doersch, C.; Pires, B.~A.; Guo, Z.~D.; Azar, M.~G.; et~al. 2020.
\newblock Bootstrap your own latent: A new approach to self-supervised learning.
\newblock \emph{arXiv preprint arXiv:2006.07733}.

\bibitem[{He et~al.(2020)He, Fan, Wu, Xie, and Girshick}]{he2020momentum}
He, K.; Fan, H.; Wu, Y.; Xie, S.; and Girshick, R. 2020.
\newblock Momentum contrast for unsupervised visual representation learning.
\newblock In \emph{Proceedings of the IEEE/CVF Conference on Computer Vision and Pattern Recognition}, 9729--9738.

\bibitem[{Holmqvist et~al.(2011)Holmqvist, Nystr{\"o}m, Andersson, Dewhurst, Jarodzka, and Van~de Weijer}]{holmqvist2011eye-tracking-guide}
Holmqvist, K.; Nystr{\"o}m, M.; Andersson, R.; Dewhurst, R.; Jarodzka, H.; and Van~de Weijer, J. 2011.
\newblock \emph{Eye tracking: A comprehensive guide to methods and measures}.
\newblock OUP Oxford.

\bibitem[{Hu(1962)}]{hu1962visual}
Hu, M.-K. 1962.
\newblock Visual pattern recognition by moment invariants.
\newblock \emph{IRE transactions on information theory}, 8(2): 179--187.

\bibitem[{Hu(2020)}]{hu2020gaze}
Hu, Z. 2020.
\newblock Gaze analysis and prediction in virtual reality.
\newblock In \emph{2020 IEEE Conference on Virtual Reality and 3D User Interfaces Abstracts and Workshops (VRW)}, 543--544. IEEE.

\bibitem[{Hu et~al.(2021{\natexlab{a}})Hu, Bulling, Li, and Wang}]{hu2021ehtask}
Hu, Z.; Bulling, A.; Li, S.; and Wang, G. 2021{\natexlab{a}}.
\newblock Ehtask: Recognizing user tasks from eye and head movements in immersive virtual reality.
\newblock \emph{IEEE Transactions on Visualization and Computer Graphics}.

\bibitem[{Hu et~al.(2021{\natexlab{b}})Hu, Bulling, Li, and Wang}]{hu2021fixationnet}
Hu, Z.; Bulling, A.; Li, S.; and Wang, G. 2021{\natexlab{b}}.
\newblock Fixationnet: Forecasting eye fixations in task-oriented virtual environments.
\newblock \emph{IEEE Transactions on Visualization and Computer Graphics}, 27(5): 2681--2690.

\bibitem[{Hu et~al.(2019)Hu, Zhang, Li, Wang, and Manocha}]{hu2019sgaze}
Hu, Z.; Zhang, C.; Li, S.; Wang, G.; and Manocha, D. 2019.
\newblock Sgaze: A data-driven eye-head coordination model for realtime gaze prediction.
\newblock \emph{IEEE transactions on visualization and computer graphics}, 25(5): 2002--2010.

\bibitem[{Johnson et~al.(2019)Johnson, Pollard, Greenbaum, Lungren, Deng, Peng, Lu, Mark, Berkowitz, and Horng}]{johnson2019mimic}
Johnson, A.~E.; Pollard, T.~J.; Greenbaum, N.~R.; Lungren, M.~P.; Deng, C.-y.; Peng, Y.; Lu, Z.; Mark, R.~G.; Berkowitz, S.~J.; and Horng, S. 2019.
\newblock MIMIC-CXR-JPG, a large publicly available database of labeled chest radiographs.
\newblock \emph{arXiv preprint arXiv:1901.07042}.

\bibitem[{Karargyris et~al.(2021)Karargyris, Kashyap, Lourentzou, Wu, Sharma, Tong, Abedin, Beymer, Mukherjee, Krupinski et~al.}]{karargyris2021creation}
Karargyris, A.; Kashyap, S.; Lourentzou, I.; Wu, J.~T.; Sharma, A.; Tong, M.; Abedin, S.; Beymer, D.; Mukherjee, V.; Krupinski, E.~A.; et~al. 2021.
\newblock Creation and validation of a chest X-ray dataset with eye-tracking and report dictation for AI development.
\newblock \emph{Scientific data}, 8(1): 1--18.

\bibitem[{Kundel et~al.(2008)Kundel, Nodine, Krupinski, and Mello-Thoms}]{kundel2008using}
Kundel, H.~L.; Nodine, C.~F.; Krupinski, E.~A.; and Mello-Thoms, C. 2008.
\newblock Using gaze-tracking data and mixture distribution analysis to support a holistic model for the detection of cancers on mammograms.
\newblock \emph{Academic radiology}, 15(7): 881--886.

\bibitem[{Le~Meur and Baccino(2013)}]{le2013methods}
Le~Meur, O.; and Baccino, T. 2013.
\newblock Methods for comparing scanpaths and saliency maps: strengths and weaknesses.
\newblock \emph{Behavior research methods}, 45(1): 251--266.

\bibitem[{Liberman and Menell(2002)}]{LauraLiberman2002BreastIR}
Liberman, L.; and Menell, J.~H. 2002.
\newblock Breast imaging reporting and data system (BI-RADS).
\newblock \emph{Radiologic Clinics of North America}, 40: 409--430.

\bibitem[{Ma et~al.(2023)Ma, Zhao, Chen, Wang, Guo, Zhang, Shen, Jiang, and Liu}]{ma2023eye}
Ma, C.; Zhao, L.; Chen, Y.; Wang, S.; Guo, L.; Zhang, T.; Shen, D.; Jiang, X.; and Liu, T. 2023.
\newblock Eye-gaze-guided vision transformer for rectifying shortcut learning.
\newblock \emph{IEEE Transactions on Medical Imaging}.

\bibitem[{Maharana(2016)}]{imagehash_survey}
Maharana, A. 2016.
\newblock \emph{Application of Digital Fingerprinting: Duplicate Image Detection}.
\newblock Ph.D. thesis.

\bibitem[{Mall, Brennan, and Mello-Thoms(2018)}]{mall2018modeling}
Mall, S.; Brennan, P.~C.; and Mello-Thoms, C. 2018.
\newblock Modeling visual search behavior of breast radiologists using a deep convolution neural network.
\newblock \emph{Journal of Medical Imaging}, 5(3): 035502.

\bibitem[{Mall, Krupinski, and Mello-Thoms(2019)}]{mall2019missed}
Mall, S.; Krupinski, E.; and Mello-Thoms, C. 2019.
\newblock Missed cancer and visual search of mammograms: what feature-based machine-learning can tell us that deep-convolution learning cannot.
\newblock In \emph{Medical Imaging 2019: Image Perception, Observer Performance, and Technology Assessment}, volume 10952, 1095216. International Society for Optics and Photonics.

\bibitem[{Manzi et~al.(2020)Manzi, Ishikawa, Di~Dio, Itakura, Kanda, Ishiguro, Massaro, and Marchetti}]{manzi2020understanding}
Manzi, F.; Ishikawa, M.; Di~Dio, C.; Itakura, S.; Kanda, T.; Ishiguro, H.; Massaro, D.; and Marchetti, A. 2020.
\newblock The understanding of congruent and incongruent referential gaze in 17-month-old infants: an eye-tracking study comparing human and robot.
\newblock \emph{Scientific Reports}, 10(1): 1--10.

\bibitem[{Matthews, Uribe-Quevedo, and Theodorou(2020)}]{matthews2020rendering}
Matthews, S.~L.; Uribe-Quevedo, A.; and Theodorou, A. 2020.
\newblock Rendering optimizations for virtual reality using eye-tracking.
\newblock In \emph{2020 22nd symposium on virtual and augmented reality (SVR)}, 398--405. IEEE.

\bibitem[{Moreira et~al.(2012)Moreira, Amaral, Domingues, Cardoso, Cardoso, and Cardoso}]{InsMoreira2012INbreastTA}
Moreira, I.; Amaral, I.; Domingues, I.; Cardoso, A.; Cardoso, M.~J.; and Cardoso, J.~S. 2012.
\newblock INbreast: toward a full-field digital mammographic database.
\newblock \emph{Academic Radiology}, 19: 236--248.

\bibitem[{Organizers(2022)}]{gazemeetml}
Organizers, G. M.~M. 2022.
\newblock NeurIPS 2022 Gaze Meets ML Workshop.

\bibitem[{Palinko et~al.(2016)Palinko, Rea, Sandini, and Sciutti}]{palinko2016robot}
Palinko, O.; Rea, F.; Sandini, G.; and Sciutti, A. 2016.
\newblock Robot reading human gaze: Why eye tracking is better than head tracking for human-robot collaboration.
\newblock In \emph{2016 IEEE/RSJ International Conference on Intelligent Robots and Systems (IROS)}, 5048--5054. IEEE.

\bibitem[{Panetta et~al.(2021)Panetta, Rajendran, Ramesh, Rao, and Agaian}]{panetta2021tufts}
Panetta, K.; Rajendran, R.; Ramesh, A.; Rao, S.~P.; and Agaian, S. 2021.
\newblock Tufts Dental Database: A Multimodal Panoramic X-ray Dataset for Benchmarking Diagnostic Systems.
\newblock \emph{IEEE Journal of Biomedical and Health Informatics}, 26(4): 1650--1659.

\bibitem[{Peng et~al.(2022)Peng, Wang, Zhu, and You}]{peng2022crafting}
Peng, X.; Wang, K.; Zhu, Z.; and You, Y. 2022.
\newblock Crafting Better Contrastive Views for Siamese Representation Learning.
\newblock \emph{arXiv preprint arXiv:2202.03278}.

\bibitem[{Radford et~al.(2021)Radford, Kim, Hallacy, Ramesh, Goh, Agarwal, Sastry, Askell, Mishkin, Clark et~al.}]{radford2021learning}
Radford, A.; Kim, J.~W.; Hallacy, C.; Ramesh, A.; Goh, G.; Agarwal, S.; Sastry, G.; Askell, A.; Mishkin, P.; Clark, J.; et~al. 2021.
\newblock Learning transferable visual models from natural language supervision.
\newblock In \emph{International Conference on Machine Learning}, 8748--8763. PMLR.

\bibitem[{Seibold et~al.(2022)Seibold, Rei{\ss}, Sarfraz, Stiefelhagen, and Kleesiek}]{seibold2022breaking}
Seibold, C.; Rei{\ss}, S.; Sarfraz, M.~S.; Stiefelhagen, R.; and Kleesiek, J. 2022.
\newblock Breaking with fixed set pathology recognition through report-guided contrastive training.
\newblock In \emph{Medical Image Computing and Computer Assisted Intervention--MICCAI 2022: 25th International Conference, Singapore, September 18--22, 2022, Proceedings, Part V}, 690--700. Springer.

\bibitem[{Selvaraju et~al.(2021)Selvaraju, Desai, Johnson, and Naik}]{selvaraju2021casting}
Selvaraju, R.~R.; Desai, K.; Johnson, J.; and Naik, N. 2021.
\newblock Casting your model: Learning to localize improves self-supervised representations.
\newblock In \emph{Proceedings of the IEEE/CVF Conference on Computer Vision and Pattern Recognition}, 11058--11067.

\bibitem[{Sowrirajan et~al.(2021)Sowrirajan, Yang, Ng, and Rajpurkar}]{sowrirajan2021moco}
Sowrirajan, H.; Yang, J.; Ng, A.~Y.; and Rajpurkar, P. 2021.
\newblock MoCo-CXR: MoCo Pretraining Improves Representation and Transferability of Chest X-ray Models.
\newblock In \emph{Proc. International Conference on Medical Imaging with Deep Learning (MIDL)}.

\bibitem[{Uppal, Kim, and Singh(2023)}]{uppal2023decoding}
Uppal, K.; Kim, J.; and Singh, S. 2023.
\newblock Decoding Attention from Gaze: A Benchmark Dataset and End-to-End Models.
\newblock In \emph{Annual Conference on Neural Information Processing Systems}, 219--240. PMLR.

\bibitem[{Valliappan et~al.(2020)Valliappan, Dai, Steinberg, He, Rogers, Ramachandran, Xu, Shojaeizadeh, Guo, Kohlhoff et~al.}]{valliappan2020accelerating}
Valliappan, N.; Dai, N.; Steinberg, E.; He, J.; Rogers, K.; Ramachandran, V.; Xu, P.; Shojaeizadeh, M.; Guo, L.; Kohlhoff, K.; et~al. 2020.
\newblock Accelerating eye movement research via accurate and affordable smartphone eye tracking.
\newblock \emph{Nature communications}, 11(1): 4553.

\bibitem[{Voisin et~al.(2013)Voisin, Pinto, Xu, Morin-Ducote, Hudson, and Tourassi}]{voisin2013investigating}
Voisin, S.; Pinto, F.; Xu, S.; Morin-Ducote, G.; Hudson, K.; and Tourassi, G.~D. 2013.
\newblock Investigating the association of eye gaze pattern and diagnostic error in mammography.
\newblock In \emph{Medical Imaging 2013: Image Perception, Observer Performance, and Technology Assessment}, volume 8673, 867302. International Society for Optics and Photonics.

\bibitem[{Vu et~al.(2021)Vu, Wang, Balachandar, Liu, Ng, and Rajpurkar}]{vu2021medaug}
Vu, Y. N.~T.; Wang, R.; Balachandar, N.; Liu, C.; Ng, A.~Y.; and Rajpurkar, P. 2021.
\newblock Medaug: Contrastive learning leveraging patient metadata improves representations for chest x-ray interpretation.
\newblock In \emph{Machine Learning for Healthcare Conference}, 755--769. PMLR.

\bibitem[{Wan et~al.(2021)Wan, Xiong, Chen, Zhang, and Wu}]{wan2021pupil}
Wan, Z.; Xiong, C.; Chen, W.; Zhang, H.; and Wu, S. 2021.
\newblock Pupil-Contour-Based Gaze Estimation With Real Pupil Axes for Head-Mounted Eye Tracking.
\newblock \emph{IEEE Transactions on Industrial Informatics}, 18(6): 3640--3650.

\bibitem[{Wang et~al.(2022)Wang, Ouyang, Liu, Wang, and Shen}]{wang2022follow}
Wang, S.; Ouyang, X.; Liu, T.; Wang, Q.; and Shen, D. 2022.
\newblock Follow My Eye: Using Gaze to Supervise Computer-Aided Diagnosis.
\newblock \emph{IEEE Transactions on Medical Imaging}.

\bibitem[{Wedel and Pieters(2017)}]{wedel2017review}
Wedel, M.; and Pieters, R. 2017.
\newblock A review of eye-tracking research in marketing.
\newblock \emph{Review of marketing research}, 123--147.

\bibitem[{Wiehe et~al.(2022)Wiehe, Schneider, Blank, Wang, Zorn, and Biemann}]{wiehe2022language}
Wiehe, A.; Schneider, F.; Blank, S.; Wang, X.; Zorn, H.-P.; and Biemann, C. 2022.
\newblock Language over Labels: Contrastive Language Supervision Exceeds Purely Label-Supervised Classification Performance on Chest X-Rays.
\newblock In \emph{Proceedings of the 2nd Conference of the Asia-Pacific Chapter of the Association for Computational Linguistics and the 12th International Joint Conference on Natural Language Processing: Student Research Workshop}, 76--83.

\end{thebibliography}

\end{document}